\documentclass{article}

\usepackage[final]{neurips_2019}

\usepackage[utf8]{inputenc} 
\usepackage[T1]{fontenc}    
\usepackage{hyperref}       
\usepackage{url}            
\usepackage{booktabs}       
\usepackage{nicefrac}       
\usepackage{microtype}      

\usepackage{amsfonts,amssymb,amsmath,amsthm}
\usepackage[UKenglish]{babel} 
\usepackage{titlesec}
\usepackage{algorithm, algpseudocode}
\usepackage{graphicx}
\usepackage{subcaption}
\usepackage{multicol, multirow}
\usepackage{adjustbox}

\usepackage{xcolor}

\graphicspath{{Figures/}}
\newlength{\figurewidth} 
\newlength{\figureheight} 

\begin{document}

\title{Efficient structure learning with automatic sparsity selection for causal graph processes}

\author{
Théophile Griveau-Billion\\
Department of Statistics\\
Imperial College London\\
180 Queen's Gate, Kensington, London SW7 2AZ, UK\\
\texttt{theophile.griveau-billion14@imperial.ac.uk}
\And
Ben Calderhead \\
Department of Statistics\\
Imperial College London\\
180 Queen's Gate, Kensington, London SW7 2AZ, UK\\
\texttt{b.calderhead@imperial.ac.uk} \\
}

\maketitle

\begin{abstract}
We propose a novel algorithm for efficiently computing a sparse directed adjacency matrix from a group of time series following a causal graph process. Our solution is scalable for both dense and sparse graphs and automatically selects the LASSO coefficient to obtain an appropriate number of edges in the adjacency matrix. Current state-of-the-art approaches rely on sparse-matrix-computation libraries to scale, and either avoid automatic selection of the LASSO penalty coefficient or rely on the prediction mean squared error, which is not directly related to the correct number of edges. Instead, we propose a cyclical coordinate descent algorithm that employs two new non-parametric error metrics to automatically select the LASSO coefficient. We demonstrate state-of-the-art performance of our algorithm on simulated stochastic block models and a real dataset of stocks from the S\&P$500$.
\end{abstract}

\section{Introduction}
Graph structures can be represented with an adjacency matrix containing the edge weights between different nodes. Since the adjacency matrix is usually sparse, common estimation techniques involve a quadratic optimization with a LASSO penalty to impose this sparsity. While the coefficient of the $L_1$ regularization term is easy to choose when we have prior knowledge of the appropriate sparsity level, in most cases we do not and the parametrization becomes non-trivial. In this paper we focus on causal relationships between time series, following the causal graph process (CGP) approach \citep{Mei:2015ig, Mei:2017db} which assumes an autoregressive model, where the coefficients of each time lag is a polynomial function of the adjacency matrix. Using the adjacency matrix instead of the graph Laplacian allows this set-up to consider directed graphs with positive and negative edge weights. \citet{Mei:2017db} simplify the problem to a combination of quadratic minimisations with $L_1$ regularization terms which they then solve with a sparse gradient projection algorithm. They choose this algorithm for its use of sparse-matrix-computation libraries, which is efficient only for highly sparse graphs; its performance deteriorates significantly with more dense graphs.

The cyclical coordinate descent algorithm is a widely used optimisation method due to its speed. Its popularity increased after \citet{Tseng:2001bd} proved its convergence for functions that decompose into a convex part and a non-differentiable but separable term, which makes it perfectly suited for solving quadratic minimizations involving an $L_1$ regularization term. In \citet{Wu:2008jd} the efficiency of both a cyclical and a greedy coordinate descent algorithm to solve the LASSO was shown. Subsequently, \citet{Wang:2013bc} demonstrated that the cyclical version is more stable and applied it to solve the graphical LASSO problem. \citet{Meinshausen:2006kb} extended the LASSO to high-dimensional graphs by performing individual LASSO regularization on each node. This approach inspired \citet{Friedman:tb} to create the graphical-LASSO with the block coordinate descent algorithm. This algorithm computes a sparse estimate of the precision matrix assuming that the data follows a multivariate Normal distribution. The use of CCD to solve this graph-LASSO problem allows for efficiently computing the sparse precision matrix in large environments. However, this algorithm focuses on simultaneous connections not causal relationships. In contrast, \citet{Shojaie:2010ei, Shojaie:2010df} proposed different approaches to compute what they define as a graph-Granger causality effect, however their models assumed prior knowledge of the structure of the underlying adjacency matrix and therefore focused on estimating the weights.

In the literature there are many proposed methods for selecting the LASSO coefficient. \citet{Banerjee:2007uy, Meinshausen:2006kb, Shojaie:2010ei, Shojaie:2010df} use a function of the probability of error, which depends on selection of a probability. \citet{Wu:2008jd, Mei:2015ig, Mei:2017db} perform an expensive grid search and use the in- and out-of-sample error to select the coefficient. \citet{Friedman:tb, Wang:2013bc} avoid the problem by simply fixing the coefficient to a selected value. These strategies either require the selection of free parameters or rely on the prediction error to find the correct LASSO coefficient.

In this work we follow the idea of the graph-LASSO of \citet{Friedman:tb} and the individual LASSO regularization of \citet{Meinshausen:2006kb} to propose a new CCD algorithm that solves the causal graph process problem of \citet{Mei:2015ig, Mei:2017db}. The CCD approach allows us to leverage the knowledge of the specific structure of the adjacency matrix to optimise the computational steps. In addition, we propose a new metric that uses the prediction quality of each node separately to select the LASSO coefficient. Our solution does not require additional parameters and produces better results than relying solely on the prediction error. Thus, our algorithm computes the directed adjacency matrix with an appropriate number of edges, as well as the polynomial coefficients of the CGP. Furthermore, the quality of the results and speed are not affected by the sparsity level of the underlying problem. Indeed, while the algorithm proposed by \citet{Mei:2015ig, Mei:2017db} scales cubically with the number of nodes, the algorithm we propose scales quadratically while automatically selecting the LASSO coefficient.

We show the performance of our solution on simulated CGPs following a stochastic block model with different sizes, levels of sparsity and history lengths. We assess the quality of the estimated adjacency matrix by considering the difference in number of edges, the percentage of correct positives and the percentage of false positives. We highlight the performance of our approach and its limits. We then run the algorithm on a financial dataset and interpret the results. Section \ref{sec:CGP} introduces signal processing on graphs. We then introduce our novel algorithm based on a coordinate descent algorithm to efficiently estimate the adjacency matrix of the causal graph process in Section \ref{sec:CGP_CCD}. In Section \ref{sec:sel_L1} we present a new non-parametric metric to automatically select the LASSO sparsity coefficient. Finally, in Section \ref{sec:applications} we present the results on simulated and real datasets.

\section{Background to signal processing on graphs}
\label{sec:CGP}
There exist many approaches for modelling a graph; in this paper we use a directed adjacency matrix $A$. An element $A_{i,j} \in \mathbb{R}$ of the adjacency matrix corresponds to the weight of the edge from node $j$ to node $i$. We consider a time dependent graph with $N$ nodes evolving over $K$ time samples. Let $x(k) \in \mathbb{R}^N$ be the vector with the value of each node at time $k$. With this formulation, the graph signal over $K$ time samples is denoted by the matrix $X(K) = [x(0) \dots x(K - 1) ] \in \mathbb{R}^{N \times K}$. We assume the graphs to follow a causal process as defined in \citet{Mei:2015ig, Mei:2017db}. The causal graph process (CGP) assumes the graph at time $k$ to follow an autoregressive process over $M$ time lags. The current state of the graph, $x(k)$, is related to the lag $l$ through a graph filter $P_l(A)$. The graph filter is considered to be a polynomial function over the adjacency matrix with coefficients $C = \{c_{l,j}\}$ defined by: $P_l(A) = \sum_{j=0}^l c_{l,j} A^j$. Without loss of generality we can fix the coefficients of the first time lag to be $(c_{1,0},c_{1,1}) = (0, 1)$. Thus, the CGP at $k$ can be expressed as:
\begin{equation}
\label{eq:CGP_eq}
x(k) =  w(k) + A x(k-1) + \dots + \left(c_{M,0}I + \dots + c_{M,M}A^M \right) x(k-M) \; .
\end{equation}
Where $w(k)$ corresponds to Gaussian noise. Hence, the problem of reconstructing the CGP from a group of time series consists of estimating the adjacency matrix $A$ and the polynomial coefficients $C$. \citet{Mei:2015ig, Mei:2017db} consider the optimisation problem:
\begin{equation}
\label{eq:min_Ac}
(A, c) = \min_{A, c} \frac{1}{2} \sum_{k=M}^{K-1} \left\|x(k) - \sum_{l=1}^M P_l(A) x(k-l)  \right\|_2^2 + \lambda_1 \| A\|_1 + \lambda_1^c \| C \|_1 \; ,
\end{equation}
where they include a LASSO penalty for both the adjacency matrix and the polynomial coefficients to enforce sparsity. They decompose this optimisation into different steps: first estimate the coefficients $R_i = P_i(A)$, then from those coefficients retrieve the adjacency matrix $A$, which allows the polynomial coefficients $C$ to be obtained via another minimisation. Due to the $L_1$ penalty, the optimisation problem is not convex for $R_1$ and $C$.

For the first step, they perform a block coordinate descent over the matrix coefficients $R_i$. Each step of the descent is quadratic except for $R_1$ which has the $L_1$ regularisation term. From the obtained matrix coefficients $R_i$ they perform an additional step in the block descent to obtain the adjacency matrix $A$. With the estimated adjacency matrix, we can reformulate the minimisation of Equation \ref{eq:min_Ac} in a function of the vector $C$ with an $L_1$ regularisation term. In their paper, they do not go into details on how they perform these minimisations and just specify that they use a sparse gradient projection algorithm. The authors argue in favour of this algorithm because it is particularly efficient, however only in the case of highly sparse problems; otherwise, for a dense graph this algorithm will scale as the cube of the number of nodes which renders it impractical. This motivates our interest in developing a novel cyclical coordinate descent (CCD) algorithm for this problem.
 
\section{Estimating the adjacency matrix with coordinate descent}
\label{sec:CGP_CCD}
The sparse gradient projection (SGP) algorithm \citet{Mei:2015ig, Mei:2017db} used to solve each step of the block coordinate descent algorithm does not take full advantage of the structure of the problem.  We therefore propose an efficient CCD algorithm for estimating the adjacency matrix and the CGP coefficients, and since the $L_1$ regularisation constraint in the optimisation problem for the matrix coefficients only applies to $R_1$, we consider the two cases $i=1$ and $i>1$ separately. The detailed steps leading to the equations presented in this section are in Appendix \ref{sec:eq_CCD}.

\subsection{Update equation for $i>1$}
In the case of $i>1$, the minimisation of Equation \ref{eq:min_Ac} on the matrix coefficient $R_i$ simplifies to a quadratic problem, which is well suited for a CCD algorithm looping over the different lags $i$. Furthermore, since the minimisation is over a matrix, CCD allows us to avoid computing a gradient over a matrix; instead we compute the update equation for matrix $R_i$ directly. We reformulate the loss function to isolate $R_i$ with $S_k^i = x(k) - \sum_{l \neq i}^M R_l x(k-l)$, thus the utility function is:
\begin{equation}
\label{eq:lagrangian_Ri}
\mathcal{L}(R_i) = \frac{1}{2} \sum_{k=M}^{K-1} \left( S_k^i - R_i x(k-i) \right)^T \left( S_k^i - R_i x(k-i) \right)  \;,
\end{equation}
where $S_k^i$, $x_k$ are vectors of size $N$. The derivative of $\mathcal{L}(R_i)$ with respect to $R_i$ is equal to zero if:
\begin{equation}
\label{eq:upd_Ri}
R_i = \left(\sum_{k=M}^{K-1} S_k^i x(k-i)^T \right) \left(\sum_{k=M}^{K-1} x(k-i) x(k-i)^T \right)^{-1} \; .
\end{equation}
which gives us an update equation for CCD. Since the matrix $\sum_{k=M}^{K-1} x(k-i) x(k-i)^T$ in equation \ref{eq:upd_Ri} is not guaranteed to be non-singular, we can perform a regularisation step by adding noise to its diagonal to compute the inverse. While this inverse step is expensive it does not depend on the matrices $R_i$. Thus, it can be computed in advance outside of the CCD loop. Hence, the update consists of a vector-vector multiplication followed by a matrix-matrix multiplication of size $N\times N$.

\subsection{CCD for $i=1$}
For $i=1$, the optimisation corresponds to Equation \ref{eq:min_Ac} with the $L_1$ regularisation term. We can follow the methodology of the previous section and derive a matrix update to compute the CCD step. However, in practice this solution produces matrices that are too dense. Hence, we instead constrain the sparsity on each node. This corresponds to running a CCD over the columns of the matrix $R_1$. Indeed, a column $j$ of the adjacency matrix corresponds to the weight of the edges going from node $j$ to the nodes it influences. To do so, we have to reformulate the loss function as a problem over the column $j$ of $R_1$. Let us denote by $R_1^{-j}$ the matrix $R_1$ without the column $j$, $R_1^j$ the column $j$ of the matrix $R_1$, $x^{-j}(k-1)$ the vector $x(k-1)$ without the term at index $j$, and $x^j(k-1)$ the value of the $j$-th term of $x(k-1)$. Thus, we can reformulate the error term to isolate the column $j$: $x(k) - \sum_{i=1}^M R_i x(k-i)  = S_k^1 - R_1^{-j} x^{-j}(k-1) - R_1^j x^j(k-1)$. Therefore, the Lagrangian of the minimisation over the column $j$ of $R_1$ with a LASSO regularisation term follows as:
\begin{equation}
\label{eq:lagrangian_R1}
\mathcal{L}(R_1^j) = \frac{1}{2} \sum_{k=M}^{K-1} \left\| S_k^1 - R_1^{-j} x^{-j}(k-1) - R_1^j x^j(k-1) \right\|_2^2 +   \lambda_1 |R_1^j | \;.
\end{equation}
Due to the non-differentiable $L_1$ term we need to use sub-gradients and thus introduce the soft-thresholding function to obtain the CCD updating equation. We define the soft-threshold function as $S(a,b) = sign(a) ( |a| - b)_+$, where $sign(a)$ is the sign of $a$ and $(y)_+ = max(0,y)$. Then, the derivative of the Lagrangian \ref{eq:lagrangian_R1} is zero if:
\begin{equation}
\label{eq:R_1j}
R_1^j = \frac{S \left( \sum_{k=M}^{K-1}\left(S_k^1 - R_1^{-j} x^{-j}(k-1) \right) x^j(k-1), \lambda_1 \right) }{ \sum_{k=M}^{K-1} (x^j(k-1))^2} \;.
\end{equation}
The CCD algorithm for $R_1$ will loop by updating each column using Equation \ref{eq:R_1j}. As for the update of $R_i$, the denominator of Equation \ref{eq:R_1j} can be computed outside the loop. Thus the complete algorithm consists of a CCD for each lag $i$ with an inner CCD loop over the columns of $R_1$. Algorithm \ref{alg:ccd_Ri} shows the complete CCD algorithm to compute the matrix coefficients $R_i$ of the CGP process. We stop the descent when the first of four criterion is reached: the maximum number of iterations is reached, the $L_1$ norm of the difference between the matrix coefficients $R$ and its previous value is below a threshold $\epsilon$, the $L_1$ norm of the difference between the new and previous in-sample MSE is below $\epsilon$ or the in-sample MSE increases. We then obtain the adjacency matrix by running an extra step of the columns CCD on $R_1$.

\begin{algorithm}
\caption{CCD algorithm to compute the matrix coefficients $R$}\label{alg:ccd_Ri}
\begin{algorithmic}[1]
\Procedure{Compute-R}{$x, M, N, K$} 
   \State Compute the denominators of Equations \ref{eq:R_1j} and \ref{eq:upd_Ri} outside the loop
   \State $R = [ zeros(N, N) \;,\; \forall i \in [1,M] ]$  \Comment{Initialise coefficients at zero}
   \While{Convergence criterion not met}
   	\State Run CCD over the columns of $R_1$ with Equation \ref{eq:R_1j}
	\State Run CCD for each lag $i>1$ with matrix update Equation \ref{eq:upd_Ri}
   \EndWhile
   \State Run the CCD over the columns of $R_1$ to obtain the adjacency matrix $A$
\EndProcedure
\end{algorithmic}
\end{algorithm}

\subsection{Retrieving the polynomial coefficients $c$}
Once we have retrieved the adjacency matrix $A$, the next step of the block coordinate descent is to estimate the polynomial coefficients $C$. To obtain these coefficients we minimise the MSE over the training set with both $L_1$ and $L_2$ regularisation terms. Hence, we can derive the CCD update for each coefficient $C_{i,j}$. We denote by $\hat{C}$ and $\hat{A}$ the estimated values of $C$ and $A$ respectively. Then, let us define $y_k = x_k - \hat{A}x_{k-1}$ and $w_k = \sum_{i',j' \neq i,j} C_{i',j'} \hat{A}^{j'} x_{k-i'}$. The error term of Equation \ref{eq:min_Ac} as a function of $C_{i,j}$ becomes $\left\| y_k - w_k - C_{i,j} \hat{A}^j x_{k-i} \right\|_2^2$. Thus, taking into account the $L_1$ and $L_2$ regularisation terms on $C$, the derivative of the Lagrangian with respect to $C_{i,j}$ is zero if:
\begin{equation}
\label{eq:Cij}
C_{i,j} = \frac{S\left( \sum_{k=M}^{K}  \left(\hat{A}^j x(k-i)  \right)^T \left(y_k - w_k \right) , (K-M) \lambda_1^c \right)}{\sum_{k=M}^{K}  \left(\hat{A}^j x(k-i)  \right)^T \left(\hat{A}^j x(k-i)\right) + 2 (K - M ) \lambda_2^c } \;.
\end{equation}
Hence, the CCD algorithm for $C$ will loop over each coefficient $C_{i,j}$ and update it with Equation \ref{eq:Cij}. This step completes the block coordinate descent to obtain the CGP process from the observed time series $x$. However, this CCD algorithm has three parameters: $\lambda_1$ for LASSO penalty used to obtain $R_1$, and $\lambda_1^c$ and $\lambda_2^c$ for the $L_1$ and $L_2$ regularisation terms used to compute the polynomial coefficients $C$. In this paper we focus on the estimation of the adjacency matrix $A$ and the choice of the LASSO parameter, while fixing the regularisation parameters of equation \ref{eq:Cij} to $\lambda_1^c=0.05$ and $\lambda_2^c=10^3$, for which the algorithm appears to be reasonably robust, as is evident from the results.

\section{Selecting the $L_1$ coefficient}
\label{sec:sel_L1}
We now introduce two new non-parametric metrics to efficiently and automatically select the LASSO coefficient, $\lambda_1$, which directly influences the sparsity of the adjacency matrix $A$. A classic approach for selecting this involves computing the cross-validation error. When applied to time series this corresponds to computing a prediction error over an out-of-sample time window, see for example \citet{Wu:2008jd, Mei:2015ig, Mei:2017db}, in which the authors use the estimated CGP variables $\hat{A}$ and $\hat{C}$ to make a prediction and use the MSE to assess its quality. This technique implies a direct relationship between the sparsity level of the adjacency matrix and the MSE of the prediction, which is questionable in practice. Figure \ref{fig:compMetrics} plots the evolution of the prediction MSE over increasing values of $\lambda_1$, alongside the percentage difference in the number of edges between the estimated adjacency matrix and the real one computed for a simulated CGP following a Stochastic Block Model (SBM) graph, i.e. the difference in number of edges divided by the total possible number of edges.

It is clear that current techniques do not produce good results and we do not want to follow \citet{Friedman:tb, Wang:2013bc} in fixing different values of $\lambda_1$ producing a sparse and a dense result without knowing which one is correct. Thus, we need a metric that weights the improvement in prediction error against an increased number of edges. In statistical modelling the AIC and BIC criteria aim to avoid over-fitting by including a penalty on the number of input-variables, which works when the number of output-variables to be predicted is not related to the selected set of input-variables. However, in the case of an adjacency matrix the sparsity of each node also impacts the number of nodes predicted; we want sparsity in the adjacency matrix to encourage each node to more accurately predict a small subset of other nodes. Following this idea further we derive two new error metrics.

\subsection{Two new distance measures for selecting the $L_1$ coefficient}
We want a distance metric that has no parameters and a maximum around the exact number of edges of the adjacency matrix. An appropriately sparse adjacency matrix has few edges but enough to accurately reproduce the whole CGP process. What is important is not the prediction quality obtained by the complete adjacency matrix but rather the prediction quality of each node {\it independently}; the adjacency matrix should have few edges connecting nodes with each connection having a low in- and out-of-sample error. For each node we compute the errors of each connection with other nodes. We sum these errors over all edges and compute the average over time divided by the number of edges. From this error, we compute the sum over all nodes to obtain an error metric of the whole graph:
\begin{equation}
\label{eq:err_j}
err = \sum_{j=1}^N \frac{1}{\sum_{i=1}^N  \mathbb{I}_{ \{A_j \neq 0 \} }(i)  } \frac{1}{K-M}\sum_{k=M}^K \left\| x(k) \mathbb{I}_{\{A_j \neq 0 \}} - A x_j(k-1) \right\|_2^2 \;,
\end{equation}
where $ 1_{A_j \neq 0} $ corresponds to a vector of zeros with ones only where the lines of the column $A_j$ of the adjacency matrix $A$ are non-zero. Compared to the in-sample MSE of node $j$ the error metric of Equation \ref{eq:err_j} focuses on the error in the nodes it is connected to. Since we divide by the number of edges, the error should increase as sparsity increases, but it should start decreasing once the gain in the prediction quality of each edge offsets the decrease in number. Intuitively this peak should correspond to the underlying sparsity level of the graph under study; before the peak, the model has too many parameters with poor individual prediction quality, whereas after the peak, the model has too few edges with very low individual error.

\begin{figure}[h]
\vspace{.3in}
\centerline{\includegraphics[width = \figurewidth, height = \figureheight]{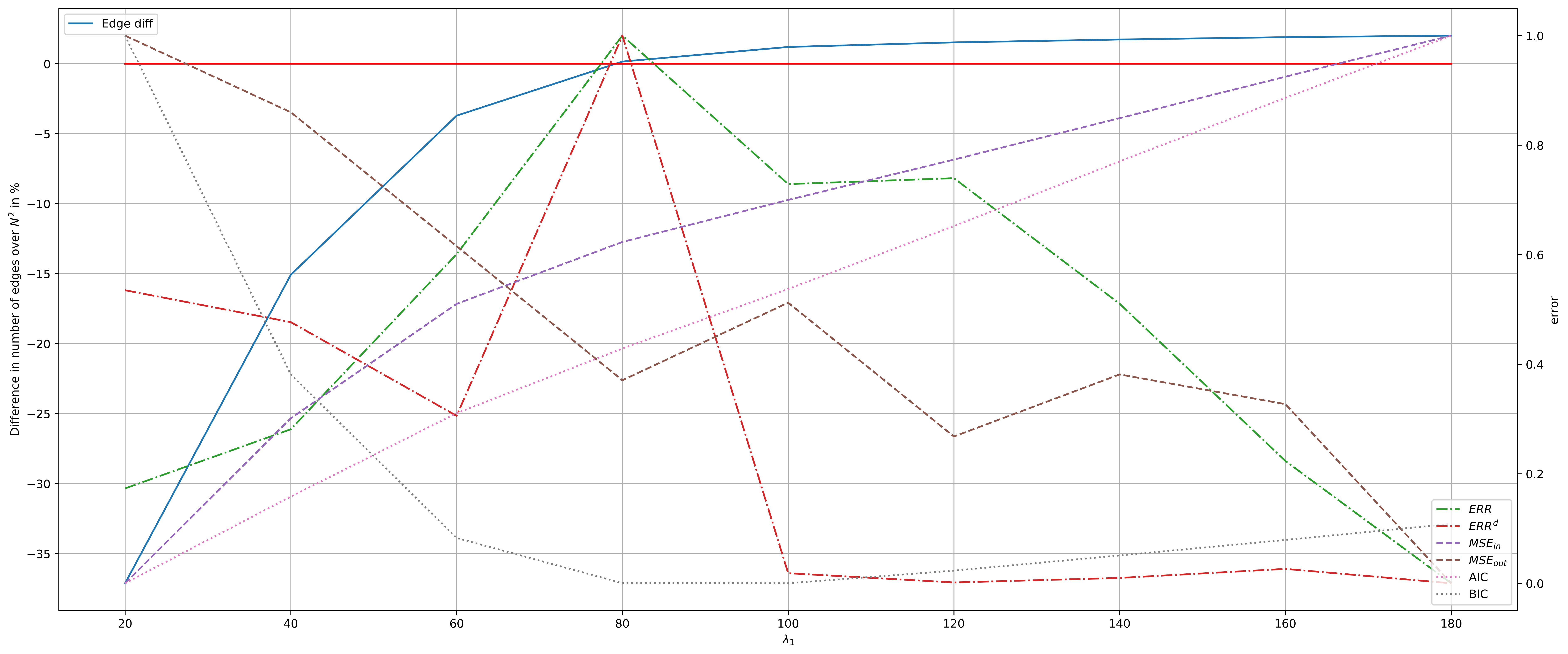}}
\vspace{.3in}
\caption{\label{fig:compMetrics} Comparison of the evolution of different metrics as a function of the value of the LASSO coefficient $\lambda_1$ for a simulated CGP-SBM graph with $200$ nodes, $5$ clusters, $3$ lags and $1040$ time points. The left axis indicates the number of different edges between the estimate and the true adjacency matrix. The blue line shows the evolution of that difference as a function of the coefficient $\lambda_1$, while the red line highlights the zero mark. The different error metrics are rescaled to be between $0$ and $1$ on the right y-axis. We compare the two metrics proposed in this paper, $err$ and $err^d$, with the in- and out-of-sample error, $MSE_{in}$ and $MSE_{out}$, as well as the AIC and BIC criteria, $AIC$ and $BIC$}
\end{figure}

The error metric of Equation \ref{eq:err_j} averages over the number of edges of each node $j$. Another approach would be to work with the degree of the graph instead, by which we mean the sum of the absolute values of the weights of the edges of node $j$. Thus we define the error with degree by:
\begin{equation}
\label{eq:err_jd}
err^d = \sum_{j=1}^N \frac{1}{\sum_{i=1}^N |A_{i,j}|  } \frac{1}{K-M}\sum_{k=M}^K \left\| x(k) \mathbb{I}_{\{A_j \neq 0 \}} - A x_j(k-1) \right\|_2^2
\end{equation}
We simulate a CGP on a stochastic block model following the methodology and parameters in \citet{Mei:2015ig, Mei:2017db}, which we describe in Section \ref{sec:applications}. On this simulated graph we test our intuition by comparing the evolution of our error metrics as a function of the sparsity parameter $\lambda_1$. Figure \ref{fig:compMetrics} shows the evolution of these metrics as well as the commonly used in- and out-of-sample MSE and the AIC and BIC criteria. On this figure both metrics, $err$ and $err^d$, perform as expected while the others do not show any relationship to the number of edges in the adjacency matrix except for the BIC criterion. While this graph represents only one sample of a simulated scenario, this behaviour is consistent throughout the different simulations in Section \ref{sec:applications}. Interestingly, our two metrics complement each other; on average, for dense graphics $err$ often produces better results while for sparser ones $err^d$ is often better. In practice we can use the following pragmatic approach for robust results: if both metrics have a peak we take the mean value of the two resulting $\lambda_1$, if only one has a peak we take its value for $\lambda_1$. The Table \ref{tab:perf_compM} in Appendix compares the performance of these different metrics for different SBMs. It is interesting to observe that the performance of the two error metrics we propose is on par with the BIC criterion and actually complement is. Indeed, we observe the best results when averaging the selected $\lambda_1$ of $err$, $err^d$ and $BIC$. Since the BIC criterion is widely known, for the rest of this paper we will study the performance of the two new error distances $err$ and $err^d$ that we propose knowing that the resulting adjacency matrix would be better if the BIC criterion was included.

\section{Applications}
\label{sec:applications}
We are interested in two performance metrics: the accuracy of the LASSO coefficient selection, assessed by measuring the difference in number of edges between the real and estimated adjacency matrix; and the quality of the CCD algorithm, assessed by measuring the percentage of true positives and false positives. For the computations we fix two parameters of our algorithm: the maximum iterations $maxIt=50$ and the convergence limit $\epsilon=0.1$. For the initialisation, we observe that the algorithm performs better when starting from zeros, i.e. $Rh=[0]$, than from random matrices. When starting with matrices of zeros, due to the block coordinate structure of the algorithm, the first step corresponds to solving the mean square problem of a CGP with only one lag, then the second step considers a CGP with two lags, whereby we learn on the errors left by the first lag, and so on. Thus, each step of the descent complements the previous lags by iteratively refining previous predictions. 

We use the same causal graph stochastic block model (CGP-SBM) structure to assess the performance of our algorithm as \citet{Mei:2017db}, and hence our results are directly comparable. However, they focus on minimising the MSE and choose the LASSO coefficient that produces the best results. Although our approach results in a higher MSE, we have shown in Section \ref{sec:sel_L1} that the MSE is not linked to the sparsity of the graph. Thus, our algorithm obtains a more accurate estimate of the adjacency matrix and its approximate sparsity.

\begin{figure}[h]
\vspace{.3in}
\centerline{\includegraphics[width = \figurewidth, height = \figureheight]{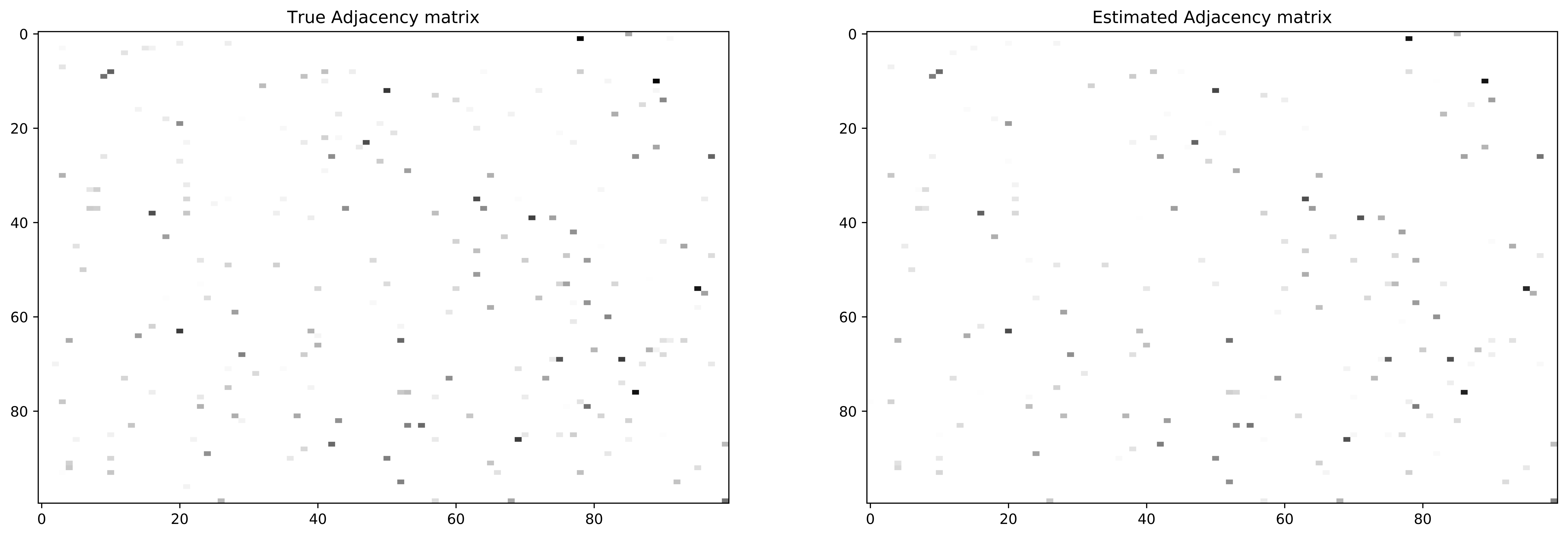}}
\vspace{.3in}
\caption{\label{fig:compA} Estimated adjacency matrix $\hat{A}$, on the right; the true one $A$, on the left. Absolute values of the weights are shown for better visualisation, with blacker points representing bigger weights. The estimation was performed on a CGP-SBM graph with $N=100$, $Nc=5$, $M=3$ and $K=1560$.}
\end{figure}

\begin{figure}[h]
\vspace{.3in}
\centerline{\includegraphics[width = \figurewidth, height = \figureheight]{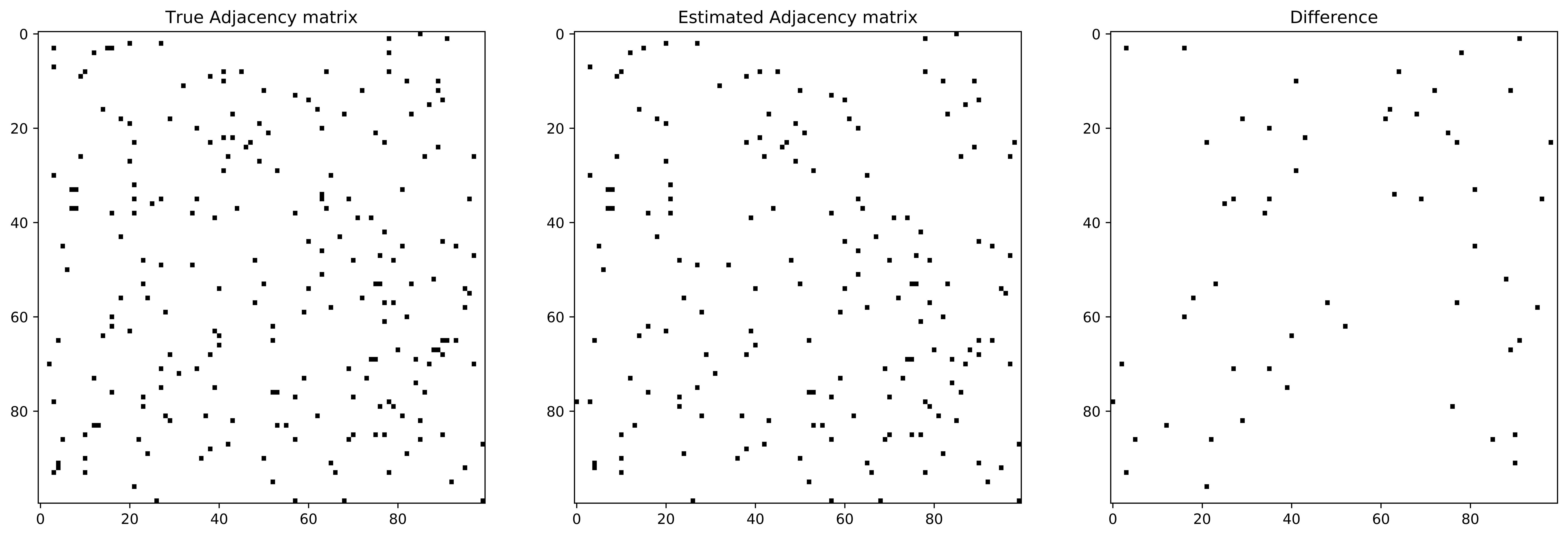}}
\vspace{.3in}
\caption{\label{fig:compAdiff_1} This figure compares the selected edges of the estimated adjacency matrix $\hat{A}$, in the middle, to the true one $A$, on the left. The right matrix correspond to the difference these two matrices. Each black square corresponds to an edge with non-zero weight. The estimation was performed on a CGP-SBM graph with $N=100$, $Nc=5$, $M=3$ and $K=1300$.}
\end{figure}

The SBM consists of a graph with a set of clusters where the probability of a connection is higher within a cluster than between them. This structure is interesting because it appears in a wide variety of real applications. The parameters to simulate a CGP-SBM  are the number of nodes, $N$, the number of clusters, $Nc$, the number of lags, $M$ and the number of time points, $K$. For each simulation we use a burn in of $500$ points. For a visual assessment of our proposed algorithm's performance we show in Figure \ref{fig:compA} the absolute value of the adjacency matrix used to simulate the CGP and the estimation we obtain with our algorithm. We note that it is hard to visually detect the discrepancies. Hence we added in appendix Figure \ref{fig:compAdiff} to show the matrix of the differences between real and estimated adjacency matrices and, Figure \ref{fig:compAdiff_1} to show the non-zero elements of each matrix and the matrix of differences with a black square at each non-zero edge.  

We also wish to quantitatively assess the accuracy of the estimated adjacency matrix. Hence, we measure the quality of the results by considering different metrics: the difference in the number of edges between $\hat{A}$ and $A$ as a absolute value and as percentage of the total number of possible edges, $N^2$; the percentage of true positive, i.e. the number of edges in $\hat{A}$ that are also edges in $A$ over the total number of edges in $A$; the percentage of false positive, i.e. the number of edges in $\hat{A}$ that are not in $A$ over the total number of edges in $\hat{A}$; the mean squared error $MSE=\|\hat{A} - A\|_2^2 / N^2$. The two metrics measuring the difference between the real and the selected number of edges assess the performance of the selection of the sparsity coefficient $\lambda_1$. The true and false positive rates assess the performance of the CGP-CCD algorithm \ref{alg:ccd_Ri} to compute the adjacency matrix.

\begin{table}[h]
\begin{center}
\caption{\label{tab:perf} Differences between the adjacency matrix $A$ and its estimate $\hat{A}$ for different CGP-SBM environments.  \textit{NBDE}: absolute difference in the number of edges between $\hat{A}$ and $A$, and as a percentage of the total number of possible edges $N^2$ with \textit{NBDE (\%)}. \textit{True positive}: number of edges in $\hat{A}$ that are edges in $A$ over the total number of edges in $A$. \textit{False positive}: number of edges in $\hat{A}$ that are not in $A$ over the total number of edges in $\hat{A}$. \textit{Mean squared error}: $MSE=\|\hat{A} - A\|_2^2 / N^2$.}

\begin{adjustbox}{max width=\textwidth}
\begin{tabular}{ c  | c | c  | c | c | c | c | c | c}
N & Nc & M & K & MSE & NBDE & NBDE (\%) & True positive & False positive \\
\hline
$100$    &  $5$     &  $3$    &  $1040$    &  $2.0\times 10^{-4}$    &  $40.5$      &  $0.41 \%$  $(0.28)$   &  $72.4 \%$ $(7.6)$   &  $20.8\%$ $(11.4)$ \\
$200$    &  $5$    &  $3$    &  $1040$    &  $2.5\times 10^{-4}$     &  $ 115.0$    &  $0.29 \%$  $(0.29)$   &  $65.9 \%$ $(4.1)$   &  $ 25.4\%$ $(9.3)$ \\
$200$    &  $10$  &  $3$    &  $1040$    &  $1.6\times 10^{-4}$     &  $198.5$     &  $0.50\%$  $(0.46)$    &  $67.0 \%$ $(8.6)$   &  $26.1 \%$ $(18.9)$ \\
$200$    &  $5$    &  $5$    &  $1040$    &  $2.3\times 10^{-4}$     &  $208.0$    &  $0.52\%$  $(0.27)$    &  $63.9 \%$ $(4.7)$   &  $26.2 \%$ $(13.7)$ \\
$200$    &  $5$    &  $3$    &  $2080$    &  $1.2\times 10^{-4}$     &  $135.5$     &  $0.34\%$  $(0.27)$    &  $73.7 \%$ $(6.3)$   &  $21.1 \%$  $(10.0)$\\
$500$    &  $5$    &  $3$    &  $2080$    &  $1.7\times 10^{-4}$     &  $1722.5$   &  $0.69 \%$  $(0.26)$   &  $ 61.3\%$ $(4.6)$   &  $17.5 \%$  $(4.9)$\\
$1000$  &  $10$   &  $3$    &  $2080$    &  $9.3\times 10^{-5}$    &  $4835.5$  &  $0.48\%$  $(0.26)$   &  $56.8 \%$ $(6.7)$   &  $17.0\%$ $(9.7)$ \\
$1000$  &  $10$   &  $3$    &  $4160$     &  $6.6\times 10^{-5}$   &  $3989.5$  &  $0.40\%$  $(0.27)$    &  $66.7 \%$ $(8.5)$   &  $15.1\%$  $(9.7)$\\
$5000$  &  $50$   &  $3$   &  $5000$    &  $1.3\times 10^{-5}$    &  $87709.5$  &  $0.35\%$   $(0.05)$  &  $64.3 \%$ $(3.5)$    &  $14.0\%$  $(8.6)$\\
\hline
\multicolumn{4}{c |}{Median} &  $1.7\times 10^{-4}$   &  $208.0$ &  $0.41\%$  $(0.12)$    &  $65.9 \%$ $(5.2)$    &  $20.8\%$ $(4.7)$ \\
\hline
\end{tabular}
\end{adjustbox}
\end{center}
\end{table}

For assessing the consistency of the performance we simulate different environments and compute the median of the measures obtained over $10$ samples. In the simulated environments, the sparsity level, i.e. the number of non-zeros elements in the adjacency matrix, varies between $1.4\%$ and $3.0\%$ with an average at $2.1\%$. Table \ref{tab:perf} shows that the results are consistent for different graph sizes, sparsity levels, lags and numbers of time points with a median over all the environments of percentage-difference in number of edges of $0.41\%$, true positive rate of $65.9\%$ and false positive rate of $20.8\%$. Interestingly, even when the number of time points was too small to obtain accurate results, i.e. high true positive rate, the percentage of different edges stays small, below $0.5\%$. The results in Table \ref{tab:perf} assume we know the number of time lags of the underlying CGP we are looking for. Since this assumption is unlikely to hold on real datasets we tested the reliability of the performances by using wrong input parameters. We therefore simulated a graph with $M=5$ lags and ran the learning algorithm for $M=3$ lags and vice-versa, in both scenarios the average results were approximately unchanged. Section \ref{sec:realData} further studies the performance of the algorithm on real datasets with unknown parameters.

\subsection{ Computation time complexity}
All the computations employed Python $2.7$ using Numpy and Scipy-sparse libraries, which can use up to $4$ threads. The Stochastic Gradient projection (SGP) method used by \citet{Mei:2015ig, Mei:2017db} is especially efficient for highly sparse environments, which can leverage sparse matrix-vector computation. In our environment, when we have a graph with more than $1\%$ of non-zero weights a matrix-matrix or matrix-vector multiplication using the sparse function of Scipy-sparse is slower than using the dense functions of Numpy. Thus, for graphs with an adjacency matrix with more than $1\%$ of non-zero edges we use the dense library. For example, on a CGP-SBM graph with parameters $(N, Nc, M, K) = (200, 5, 3, 2080)$ with $2,4\%$ of non-zeros edges, our CCD algorithm solves the optimisation of Equation \ref{eq:lagrangian_R1} to obtain the matrix $R_1$ faster than the SGP algorithm by more than $100$-fold. When the graph size increases to $N=500$, CCD is faster than by more than $350$-fold.

We perform an empirical time complexity estimation of the complete block coordinate descent algorithm by measuring the evolution of the execution time as a function of each parameter $(N, Nc, M, K)$ individually. Increasing the number of time lags $M$ has a negligible effect on the execution time of the CCD to obtain the adjacency matrix, although the computation of the polynomial coefficients $C$ scales quadratically with the lags $M$. We observe that the execution time is not affected by the sparsity level $Nc$. However, it scales linearly as function of the number of time points $K$ and quadratically as a function of the number of nodes $N$. This quadratic complexity in $N$ can be tempered in different ways however. In the case of highly sparse graphs we can leverage sparse libraries, and an even faster solution for both dense and sparse graph is to perform the computations on a GPU. Indeed, Algorithm \ref{alg:ccd_Ri} does not use much memory and can thus be computed entirely in the GPU memory. With a GPU implementation using the library PyTorch, the algorithm has a $20$-fold speed-up compared to our CPU implementation with matrix computations parallelised over $4$ threads.

\subsection{An application to financial time series}
\label{sec:realData}
We now apply our algorithm to a real dataset of stock prices consisting of the $371$ stocks from the S\&P$500$ that have quotes between $2000/01/03$ and $2018/03/27$. Since we do not know the exact adjacency matrix of this environment, we test the accuracy of the obtained graph by studying how it changed following a known market shock. More specifically we compute two graphs, one before and one after the financial crisis of $2008/2009$. For both graphs to use the same number of time points, the first uses prices from $2004/11/15$ to $2009/01/01$ and the second from $2009/11/12$ to $2014/01/01$. We chose to build the graphs with a 4 year time window, $K=1040$, since in simulations with the same number of nodes it produces good results. The lag was fixed to one week, $M=5$, since there are documented trading patterns at a weekly frequency. We note that in the time windows studied modifying the lags to $M=3$ or $M=10$ has negligible impact on the results.

For both time windows the error metrics peaked at slightly different values, thus we took the mean of the two for estimating the LASSO coefficient. Interestingly, the algorithm selects a much sparser matrix after the crisis with a sparsity level decreasing from $5.1\%$ to $2.8\%$. This points to a more inter-connected market leading up to the crisis. Since the crisis was due to sub-prime issues, one might expect real-estate and financial firms to have many edges in the graph and influence the market in the pre-crisis period, with the importance of these firms decreasing after the crash. Indeed, this aspect is reflected in the estimated adjacency matrix; before the crisis, financial firms represent more than $60\%$ of the top ten nodes with the highest number of connections, while it decreases to less than $40\%$ afterwards, including insurance firms. Furthermore, before the crisis the firm with the highest number of connection was GGP Inc., a real estate firm which went on to file for bankruptcy in $2009$. While financial and oil firms represented more than $70\%$ of the top $20$ most connected nodes before the crisis, the graph of $2014$ is much more diversified with more sectors in the top $20$ and none representing more than $30\%$. Figure \ref{fig:US_A} in the appendix shows the evolution of the adjacency matrix before and after the crisis, and we can see the shift in importance and the increase in sparsity. Overall, the post-crisis market is sparser and less concentrated than before $2009$, with fewer edges linked to financial firms.

\begin{figure}[h]
\vspace{.3in}
\centerline{\includegraphics[width=\figurewidth, height=1.1\figureheight]{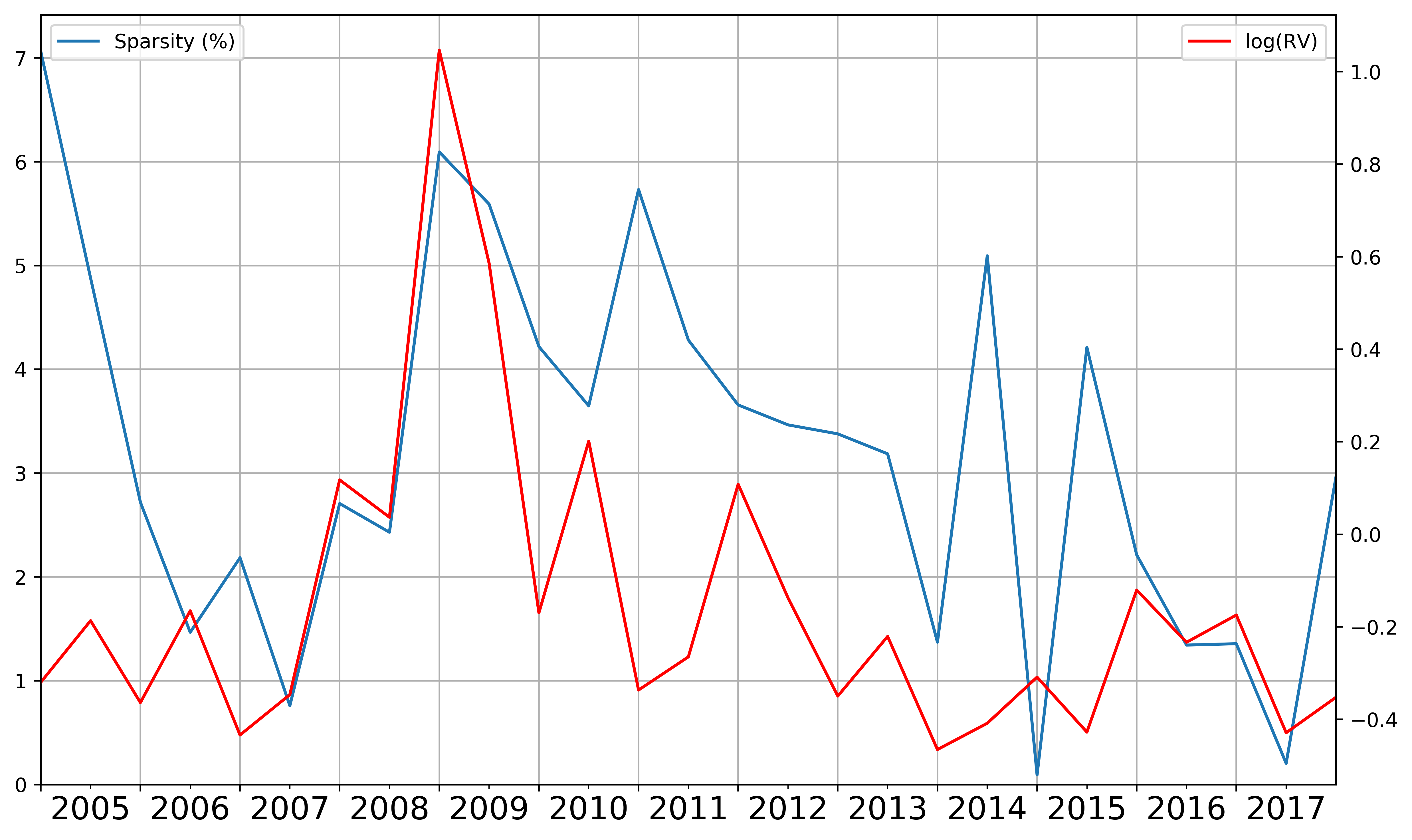}}
\vspace{.3in}
\caption{\label{fig:US_RV} For $371$ stocks from the S\&P$500$, evolution of the sparsity level of the adjacency matrix,
left y-axis, and the $log(RV)$ on the right y-axis. The coefficient $\lambda_1$ value obtained with the error metrics $err$ and $err^d$. We define the sparsity level by the percentage of non-zeros edges, hence $100.0$ corresponds to a fully connected graph. We estimate the variance of the market by the average of the log-Realised-Variance, $log(RV)$, of each stock. We computed the $log(RV)$ with exponential weighting on the daily log-returns, using a coefficient of $0.99$ and time window of $40$ days.}
\end{figure}

Since the CGP-CCD algorithm automatically selects the sparsity level, in addition to studying the different connections we can also study the evolution of the sparsity level. Figure \ref{fig:US_RV} shows the evolution through time of the sparsity level of the adjacency matrix and of the log realised variance, $log(RV)$, of the market. We computed the adjacency matrix every $6$ months and the corresponding $log(RV)$ at the last date of that time window. We can observe an interesting correlation between the increase in density of the causal graph and the increase in the realised variance.

\section{Conclusion}
We have proposed a novel cyclical coordinate descent algorithm to efficiently infer the directed adjacency matrix and polynomial coefficients of a causal graph process. Compared to the previous state-of-the-art our solution has lower complexity and does not depend on the sparsity level of the graph for scalability. Furthermore, we propose two new error metrics to automatically select the coefficient of the LASSO constraint. Our solution is able to recover approximately the correct number of edges in the directed adjacency matrix of the CGP. The performance of our algorithm is consistent across the different simulated stochastic block model graphs we tested. In addition, we provided an example application to a real-world dataset consisting of stocks from the S\&P$500$, demonstrating results that are in line with economic theory.

\bibliography{bibTexLibGraph}{}
\bibliographystyle{humannat}

\section{SUPPLEMENTARY MATERIAL}
\subsection{Detailed derivation of the CCD equations} \label{sec:eq_CCD}
\subsubsection{For $i>1$}
Recall the Utility function for $i>1$ detailed in Equation \ref{eq:lagrangian_Ri}:
\begin{equation*}
\mathcal{L}(R_i) = \frac{1}{2} \sum_{k=M}^{K-1} \left( S_k^i - R_i x(k-i) \right)^T \left( S_k^i - R_i x(k-i) \right)  \;,
\end{equation*}
With $S_k^i = x(k) - \sum_{l \neq i}^M R_l x(k-l)$. Then, the derivative of this function with respect to the matrix $R_i$ is:
\begin{equation*}
    \frac{\partial \mathcal{L}(R_i)}{\partial R_i} = R_i \sum_{k=M}^{K-1} x(k-i) x(k-i)^T - \sum_{k=M}^{K-1} S_k x(k-i)^T \;.
\end{equation*}{}
This derivative is equal to zero if $R_i$ follows the Equation \ref{eq:upd_Ri}:
\begin{equation*}
    R_i = \left(\sum_{k=M}^{K-1} S_k^i x(k-i)^T \right) \left(\sum_{k=M}^{K-1} x(k-i) x(k-i)^T \right)^{-1} \; .
\end{equation*}{}
However, the matrix denominator $\sum_{k=M}^{K-1} x(k-i) x(k-i)^T$ is not guaranteed to be positive semi-definite. In case of singularity we can add an $L_2$ regularisation term, $\lambda_2 \|R_i \|_2^2$, to the utility function of Equation \ref{eq:lagrangian_Ri}. With this coefficient the updating equation of matrix $R_i$ becomes:
\begin{equation*}
    R_i = \left(\sum_{k=M}^{K-1} S_k^i x(k-i)^T \right) \left(\sum_{k=M}^{K-1} x(k-i) x(k-i)^T + 2 \lambda_2 \mathbb{I}_N \right)^{-1} \; .
\end{equation*}{}
With $\mathbb{I}_N$ the identity matrix of size $N$. Thus, with this coefficient we add noise to the diagonal of the matrix to make it non-singular. In practice, Since this coefficient adds a bias, we select the lowest value of the coefficient $\lambda_2$ that makes this matrix non-singular.

\subsubsection{For $i=1$}
In the case of $i=1$ we apply the CCD algorithm by iterating over the columns, let us recall the Lagrangian detailed in Equation \ref{eq:lagrangian_R1}:
\begin{equation*}
\mathcal{L}(R_1^j) = \frac{1}{2} \sum_{k=M}^{K-1} \left\| S_k^1 - R_1^{-j} x^{-j}(k-1) - R_1^j x^j(k-1) \right\|_2^2 +   \lambda_1 |R_1^j | \;.
\end{equation*}
which is a non-differentiable function due to the LASSO regularisation term. Thus we compute the sub-gradient of the Lagrangian with respect to the vector of the column $R_1^j$:
\begin{equation*}
    \frac{\partial \mathcal{L}(R_1^j)}{\partial R_1^j} = R_1^j \sum_{k=M}^{K-1} \left(x^j(k-1)\right)^2 - \sum_{k=M}^{K-1} \left[S_k^1 - R_1^{-j} x^{-j}(k-1) \right] x^j(k-1)^T + \lambda_1 \Gamma_j \;.
\end{equation*}{}
Where $\Gamma_j(i)=sign(R_1^j(i))$ if $R_1^j(i) \neq 0$, else $\Gamma_j(i) \in [-1, 1]$ if $R_1^j(i) = 0$. Which is why we introduce the soft-thresholding function $S(a,b) = sign(a) ( |a| - b)_+$, where $sign(a)$ is the sign of $a$ and $(y)_+ = max(0,y)$. Hence, the derivative is equal to zero if $R_1^j$ follows Equation \ref{eq:R_1j}:
\begin{equation*}
R_1^j = \frac{S \left( \sum_{k=M}^{K-1}\left(S_k^1 - R_1^{-j} x^{-j}(k-1) \right) x^j(k-1), \lambda_1 \right) }{ \sum_{k=M}^{K-1} (x^j(k-1))^2} \;.
\end{equation*}

\subsection{Figures \& Tables}
In Table \ref{tab:perf_compM}, in order to compare the different distance measure to select the coefficient $\lambda_1$ we ran the computations for a simulated CGP-SBM with the same performance metrics as in Table $1$.  We used a grid of $\lambda_1 \in [30, 300]$ with a step of $5$; in this window the AIC and $MSE_{in}$ did not converge, hence we report the result obtained for the smallest value $\lambda_1=30$.  While on a small graph, with $200$ nodes, the BIC has results on par with our distance measures $err$ \& $err^d$, for a larger problem with $500$ nodes the difference in performance becomes more significant. In these experiments, the BIC criteria is smoother than $err$ \& $err^d$ but often overestimates the number of connections compared to $err$ \& $err^d$. Thus, the average of those three metrics $err$ \& $err^d$ \& BIC gives results for the NBDE with lower variance than the others, for a small decrease in accuracy compare to  $err$ \& $err^d$ alone.
\begin{table}[h]
\begin{center}
\caption{\label{tab:perf_compM} Differences between $A$ and its estimate $\hat{A}$ on CGP-SBM graph for different distance metrics.}

\begin{adjustbox}{max width=\textwidth}
\begin{tabular}{ l | c c c c | c  c  c  c  c}
 Distance measure & N & Nc & M & K & NBDE & NBDE (\%) & True positive & False positive \\
\hline
$Err$ \& $Err^d$ & 200 & 5 & 3 & 1040     &  $ 242$    &  $0.39 \%$  $(0.35)$   &  $62.7 \%$ $(6.0)$   &  $ 15.4\%$ $(9.3)$ \\
BIC   		     & 200 & 5 & 3 & 1040     &  $175$      &  $0.48 \%$  $(0.20)$   &  $62.8\%$ $(4.1)$   &  $21.8\%$ $(5.8)$ \\
$MSE_{out}$  	 & 200 & 5 & 3 & 1040     &  $615$        &  $1.54 \%$  $(0.35)$    &  $40.3\%$ $(11.6)$   &  $0.2\%$ $(3.2)$ \\
AIC 			 & 200 & 5 & 3 & 1040     &  $8124$      &  $20.31 \%$  $(5.36)$   &  $82.2\%$ $(4.2)$   &  $91.0\%$ $(3.5)$ \\
$MSE_{in}$  	 & 200 & 5 & 3 & 1040     &  $8124$      &  $20.31\%$  $(5.36)$   &  $82.2\%$ $(4.2)$   &  $91.0\%$ $(3.5)$ \\
\hline
$Err$ \& $Err^d$        & 500 & 5 & 3 & 2080     &  $785$     &  $0.31 \%$  $(0.33)$   &  $63.0 \%$ $(7.0)$   &  $ 23.9\%$ $(10.4)$ \\
BIC   		            & 500 & 5 & 3 & 2080     &  $1552$    &  $0.62 \%$  $(0.33)$   &  $74.3\%$ $(4.5)$   &  $40.8\%$ $(9.0)$ \\
$Err$ \& $Err^d$ \& BIC & 500 & 5 & 3 & 2080     &  $942$     &  $0.38\%$  $(0.28)$   &  $66.8\%$ $(5.5)$   &  $27.1\%$ $(10.9)$ \\
\hline
$Err$ \& $Err^d$        & 1000 & 10 & 3 & 2080     &  $4059$    &  $0.33 \%$  $(0.34)$   &  $64.9 \%$ $(8.9)$   &  $ 18.8\%$ $(9.3)$ \\
BIC   		            & 1000 & 10 & 3 & 2080     &  $3106$    &  $0.31 \%$  $(0.21)$   &  $71.3\%$ $(6.9)$   &  $35.8\%$ $(9.2)$ \\
$Err$ \& $Err^d$ \& BIC & 1000 & 10 & 3 & 2080     &  $1965$    &  $0.16\%$  $(0.33)$    &  $67.8\%$ $(8.1)$   &  $24.8\%$ $(9.9)$ \\
\end{tabular}
\end{adjustbox}
\end{center}
\end{table}

\begin{figure}[h]
\vspace{.3in}
\centerline{\includegraphics[width = \figurewidth, height = \figureheight]{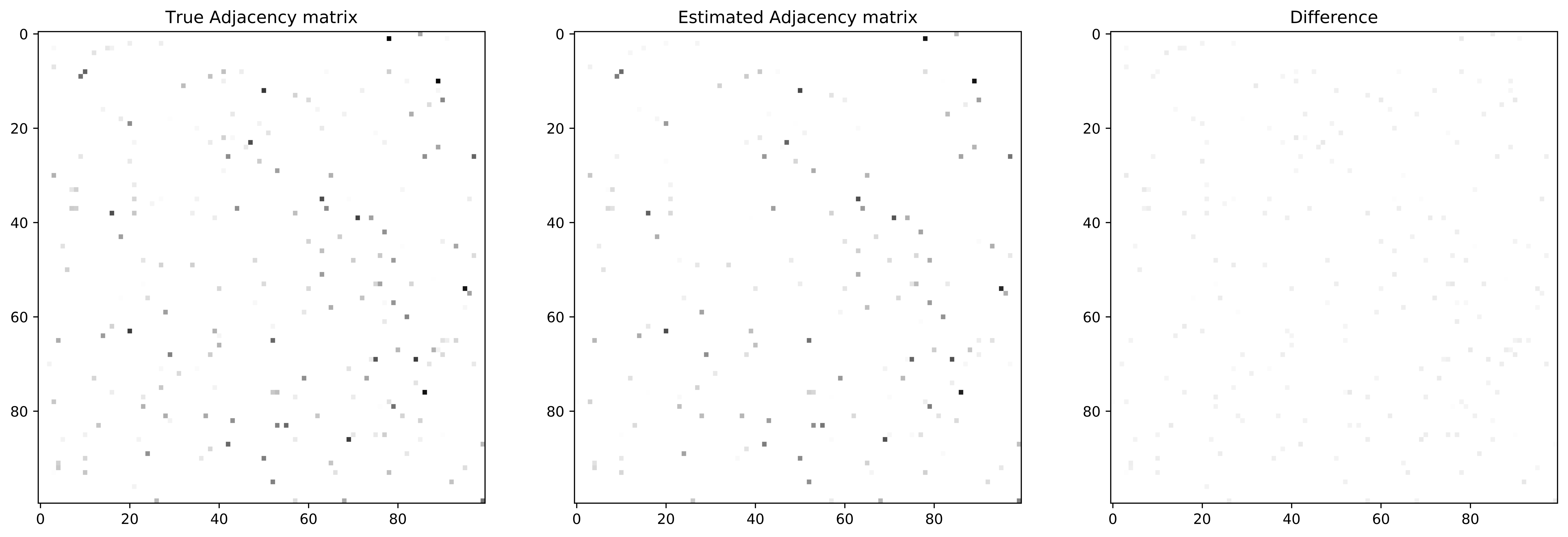}}
\vspace{.3in}
\caption{\label{fig:compAdiff} This figure compares the estimated adjacency matrix $\hat{A}$, on the right, to the true one $A$, on the left. The right matrix correspond to the difference these two matrices. To obtain a better visualisation the plots represent the absolute values of the weights, hence the blacker the bigger the weight. The estimation was performed on a CGP-SBM graph with $N=100$, $Nc=5$, $M=3$ and $K=1300$.}
\end{figure} 



\begin{figure}[h]
\vspace{.3in}
\centerline{\includegraphics[width=\figurewidth, height=1.1\figureheight]{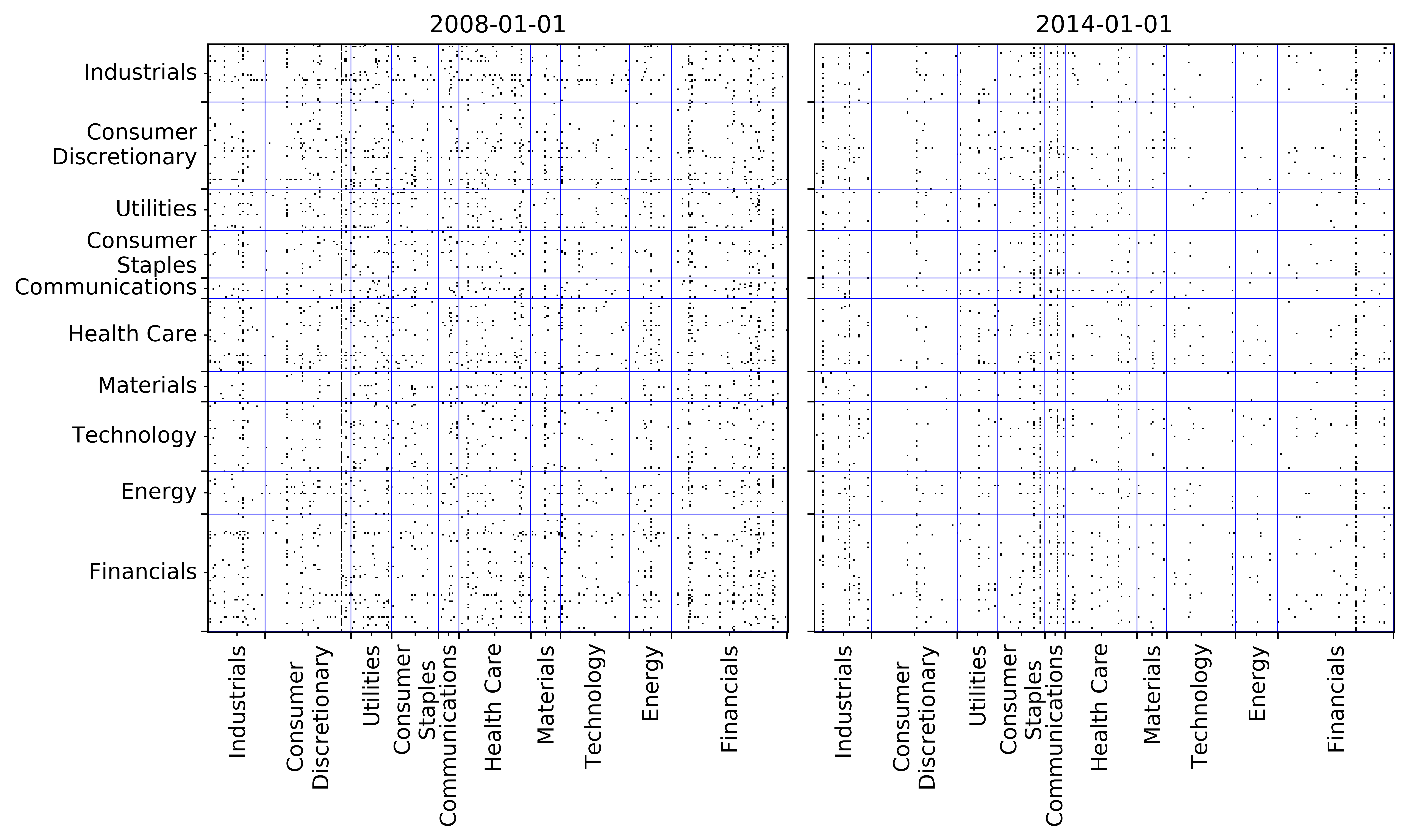}}
\vspace{.3in}
\caption{\label{fig:US_A} This figures shows the obtained adjacency matrix for a weekly lag $M=5$ on $467$ European stocks computed using data between $06/11/2003$ and $01/01/2008$ on the left, and $03/11/2009$ to $01/01/2014$ on the right. The stocks are grouped by financial sectors using the BCIS (Bloomberg Industry Classification Standard) sector classification. For a better visualisation we do not show the weight of the edge but instead a black square for every non-zero edge. The sparsity level was selected by taking the average $\lambda_1$ value obtained with the error metrics $err$ and $err^d$.}
\end{figure}

\end{document}